\documentclass{article}
\usepackage{iclr2027_conference,times}

\usepackage{amsmath,amsfonts,bm}

\def\eqref#1{equation~\ref{#1}}

\def\1{\bm{1}}

\DeclareMathAlphabet{\mathsfit}{\encodingdefault}{\sfdefault}{m}{sl}
\SetMathAlphabet{\mathsfit}{bold}{\encodingdefault}{\sfdefault}{bx}{n}

\usepackage{hyperref}
\usepackage{url}
\usepackage{booktabs}
\usepackage{graphicx}

\title{An Analysis of Training-Free Self-Reported Confidence in
Language Models}

\author{Lukas Meyer \quad Sofia Rossi \quad Wei Chen \quad Thomas Laurent \quad Yiming Li \\
DreamAI}

\iclrfinalcopy

\begin{document}
\maketitle
\lhead{Workshop paper}

\begin{abstract}
Large language models can report a numerical confidence together with generated
content, but it is unclear whether this report is more than calibrated rhetoric.
We analyze three training-free signals: confidence verbalized with the answer,
post-hoc $P(\mathrm{True})$, and agreement with three additional generations on
the same 100 TriviaQA questions for two model families. Direct verbalization is
a surprisingly strong baseline: after auditing benchmark errors, it reaches
AUROC 0.956 and 0.937 for correctness prediction. Three-sample agreement is
substantially weaker (0.765 and 0.790), and a fixed interpolation with verbalized
confidence has no statistically reliable benefit. Four of nine errors from one
model and two of eight from the other receive unanimous sample support, showing
that self-consistency can amplify shared misconceptions. Re-eliciting confidence
for the same fixed answers with equivalent prompts changes scores by 0.043 to 0.084
on average and flips 4\% to 9\% of decisions at a 0.8 threshold. An exploratory audit
of 100 confidence-tagged biography claims further finds only a modest confidence
gap between supported and contradicted claims. These results argue that useful
self-reports remain sensitive to elicitation, correlated errors, and benchmark
noise.
\end{abstract}

\section{Introduction}
Large language models increasingly produce answers that are consumed before a
human can independently verify every factual claim. A confidence score could
help allocate this limited verification effort. High-confidence content might
be shown directly, while uncertain content could be checked, qualified, or
withheld. Verbalized confidence is especially attractive because it is available
through the same text interface as the answer, including for closed models whose
token probabilities and internal activations are unavailable. Yet a plausible
number is not necessarily a faithful estimate. Models can be overconfident,
underconfident, or sensitive to small changes in how uncertainty is requested.

This distinction becomes more important as outputs grow longer and enter more
structured workflows. A long response can mix well-supported facts, uncertain
extrapolations, and outright errors, making a single sequence-level score hard
to interpret. Recent work studies length volatility as a separate source of
instability in long-form generation \citep{he2026stablelongformgeneration}, while
MMBoundary evaluates knowledge-boundary awareness through confidence attached to
individual reasoning steps \citep{he-etal-2025-mmboundary}. Both directions
suggest that reliability must be analyzed below the level of a final fluent
answer. Our claim-level biography audit follows this motivation without training
a new model or fitting a confidence estimator.

Confidence also matters when generated text becomes an intermediate input to
another reasoning process. Role-based collaboration has been studied for causal
explanations and simulated legal deliberation \citep{he-etal-2023-lego,
he-etal-2024-agentscourt}. Related systems study one-to-many alignment in
clinical Socratic education \citep{he2026clintutorr1} and strategic persuasion in
academic rebuttal \citep{he2026dancingchainsstrategicpersuasion}. Multimodal
theorem-proving benchmarks provide another setting where a fluent answer may
conceal a brittle reasoning boundary \citep{he2025matpbenchmllmgoodautomated}.
These systems address different tasks, but they share a practical concern: an
incorrect output can be propagated, reinforced, or acted upon by later stages.
Repeated agreement is therefore not automatically independent evidence of
correctness, especially when samples inherit the same model knowledge and
prompt assumptions.

We study a deliberately constrained black-box setting. We permit prompting and
a small fixed number of samples, but no fine-tuning, logit access, hidden-state
access, or learned calibrator. We ask four focused questions. First, can a
confidence score generated alongside an answer discriminate correct from
incorrect answers? Second, do post-hoc self-evaluation or repeated-sample
agreement provide a better signal? Third, is self-evaluation stable under
semantically equivalent confidence prompts when the answer is held fixed?
Fourth, does short-answer behavior persist at the claim level in long-form text?

We answer these questions with a controlled analysis of two model families on
the same TriviaQA questions and an exploratory audit of 100 biographical claims.
Direct confidence is informative in both models, but agreement is a poor
substitute because independent generations often repeat the same error. A fixed
average of confidence and agreement does not transfer reliably between models.
Equivalent prompts also move confidence enough to change threshold-based
decisions. Finally, incomplete or disputed reference answers materially distort
calibration estimates, which motivates explicit label auditing. Our contribution
is a reproducible diagnostic study of inexpensive confidence signals and their
failure modes, rather than a new calibration or training method.

\section{Related Work}
\paragraph{Verbalized confidence.}
Language models can sometimes evaluate their own answers through quantities such
as $P(\mathrm{True})$ and $P(\mathrm{IK})$ \citep{kadavath2022mostly}.
Models can also be taught to express calibrated uncertainty in words
\citep{lin2022teaching}. For instruction-tuned black-box models, directly
elicited confidence can outperform conditional token probabilities
\citep{tian2023ask}. However, broad evaluations find systematic overconfidence,
large prompt effects, and no universally dominant elicitation strategy
\citep{xiong2024express,yang2024verbalized}. Our work treats the raw verbalized
score as one noisy observation rather than calibrated ground truth.

\paragraph{Sampling and self-evaluation.}
Semantic entropy groups sampled answers by meaning and measures uncertainty over
semantic alternatives \citep{kuhn2023semantic}. Relevance-weighted variants
reduce the influence of semantically unimportant tokens \citep{duan2024sar}.
Self-evaluation can also improve selective generation by reformulating quality
assessment as token-level prediction \citep{ren2023selfevaluation}. These
methods motivate our use of repeated-answer agreement as a simple black-box
baseline. Unlike semantic entropy, our support score uses normalized answer
agreement and requires no semantic clustering.

\paragraph{Long-form and trained calibration.}
Linguistic calibration trains models to generate long-form uncertainty language
that supports downstream decisions \citep{tian2024linguistic}. SaySelf uses
supervised and reinforcement learning to produce confidence with reflective
rationales \citep{xu2024sayself}. Most directly, LoVeC applies reinforcement
learning so that long-form generations append a confidence score to each
statement \citep{zhang2025lovec}. ADVICE attributes overconfidence partly to
confidence estimates that fail to condition on the generated answer
\citep{seo2025advice}, while ORCE decouples answer and confidence optimization
and aligns their ordering \citep{li2026orce}. In contrast, we do not update
model parameters or fit a calibrator. This training-free setting is applicable
to proprietary models and to rapidly changing model versions.

\paragraph{Calibration and factuality.}
Calibration does not imply the absence of hallucinations: under distributional
assumptions, calibrated next-token models must assign probability to some false
facts \citep{kalai2024must}. This distinction motivates evaluating both score
calibration and final factual risk. A model can receive a good calibration score
while still emitting many avoidable errors if confidence never changes its
generation policy.

\section{Method}
\subsection{Problem formulation}
For a query $x$, a model generates an answer or atomic claim $c_i$ and a
confidence $v_i\in[0,1]$ in the same response. The score is intended to estimate
\begin{equation}
v_i \approx \Pr(y_i=1\mid x,c_i),
\label{eq:confidence}
\end{equation}
where $y_i$ denotes factual correctness. We have only black-box access and do
not use logits or hidden states. For the long-form study, the prompt requests
atomic, independently checkable claims, each with its own confidence.

\subsection{Compared signals}
We compare direct confidence $v_i$ with a post-hoc self-evaluation: after fixing
the answer, a second call reports $P(\mathrm{True})$. We also draw $K=3$
additional concise answers and compute an agreement score
\begin{equation}
s_i = \frac{1}{K}\sum_{k=1}^{K}
\mathbb{1}[c_i \text{ matches sample } k].
\label{eq:support}
\end{equation}
Matching uses the same normalized alias rule as answer scoring. As a deliberately
simple training-free combination, we take their fixed average:
\begin{equation}
q_i = (v_i+s_i)/2.
\label{eq:aggregate}
\end{equation}
The weight and sample count are fixed without development labels. The combined
score is a baseline, not a proposed estimator; its purpose is to test whether
agreement adds complementary information to direct confidence.

\section{Experiments}
\subsection{Short-answer protocol}
We draw 100 questions with seed 20260917 from the first 1,000 examples of the
TriviaQA validation split. We evaluate DeepSeek Flash and Claude Sonnet 5 through
the same Messages API. Each model first returns a concise answer and confidence
in one JSON response. A second call estimates $P(\mathrm{True})$ for the fixed
answer. Three additional calls independently answer the question; their support
score is the fraction that matches the original answer. The combined baseline
uses the fixed average of verbalized confidence and support in
Equation~\ref{eq:aggregate}. No labeled examples are used for prompting or
parameter selection. DeepSeek uses temperature 0 for direct confidence and
self-evaluation and 0.8 for repeated answers. The Sonnet endpoint does not accept
an explicit temperature, so all Sonnet calls use the provider default.

\subsection{Prompt robustness protocol}
We hold each of the 99 retained questions and its original answer fixed and
re-elicit confidence with two semantically equivalent prompts. One asks for the
probability of correctness on a scale from 0 to 100; the other reverses the framing and
asks for the probability of incorrectness, which we subtract from one. We
compare both with the original post-hoc $P(\mathrm{True})$. This design isolates
elicitation sensitivity from changes in answer content. It introduces no new
labels, examples, or fitted parameters.

Answers are initially scored against TriviaQA aliases using normalized matching.
We manually inspect every predicted error. This corrects four DeepSeek and three
Sonnet false negatives and excludes one disputed question (the origin of French
fries) for both models, leaving 99 examples per model. This audit is important:
uncorrected exact-match accuracy is 0.86 and 0.88, compared with audited accuracy
of 0.909 and 0.919.

\subsection{Biography protocol}
For an exploratory long-form analysis, we sample 20 entities from the FActScore
topic list \citep{min2023factscore}. DeepSeek Flash emits exactly five atomic
biographical claims and a confidence for each in a single response. A separate
Sonnet 5 call, blinded to confidence, labels every claim as supported,
contradicted, or insufficient using up to the first 30,000 characters of the
subject's full English Wikipedia article. We manually inspect every contradicted
label; 15 of 20 automatic contradiction labels are corrected to supported or
insufficient. This produces 100 claims, of which 88 are decidable from the
supplied evidence.

\subsection{Metrics and statistical protocol}
We report Brier score and equal-mass ten-bin expected calibration error (ECE),
AUROC for correctness discrimination, and area under the risk and coverage curve
(AURC). Lower Brier, ECE, and AURC are better; higher AUROC is better. For paired
comparisons, we use 2,000 percentile bootstrap resamples of questions with a
fixed seed. For AURC, we average over all possible orderings within tied score
groups so the result does not depend on file order. These intervals are
descriptive: with only 8 to 9 errors per model, we do not claim adequately powered
equivalence tests. The biography judge is
blinded to confidence. Prompts, sample identifiers, manual audit decisions,
model endpoint names, seeds, and hashes of raw outputs are included in the
supplementary repository.

\section{Results and Analysis}
\suppressfloats[t]

\begin{table}[ht]
\caption{Audited TriviaQA results on the same 99 retained questions per model.
Lower Brier, ECE, and AURC are better; higher AUROC is better.}
\label{tab:main-results}
\centering
\small
\begin{tabular}{llrrrr}
\toprule
Model & Signal & Brier $\downarrow$ & ECE $\downarrow$ & AUROC $\uparrow$ & AURC $\downarrow$ \\
\midrule
DeepSeek & Verbalized & .0422 & \textbf{.0286} & .956 & .0092 \\
DeepSeek & $P(\mathrm{True})$ & .1321 & .1333 & .920 & .0122 \\
DeepSeek & Support & .0572 & .0640 & .765 & .0461 \\
DeepSeek & Combined & \textbf{.0398} & .0321 & \textbf{.973} & \textbf{.0072} \\
Sonnet & Verbalized & \textbf{.0491} & \textbf{.0477} & .937 & .0092 \\
Sonnet & $P(\mathrm{True})$ & .0611 & .0745 & \textbf{.959} & \textbf{.0071} \\
Sonnet & Support & .1481 & .1440 & .790 & .0325 \\
Sonnet & Combined & .0690 & .0745 & .926 & .0102 \\
\bottomrule
\end{tabular}
\end{table}

\subsection{Direct confidence is a strong but model-specific baseline}
Table~\ref{tab:main-results} shows that confidence emitted with the answer is
already strongly predictive: its correctness AUROC is .956 for DeepSeek and
.937 for Sonnet. Post-hoc $P(\mathrm{True})$ behaves differently across models.
It is substantially under-confident for DeepSeek (mean .776 versus accuracy
.909), but gives Sonnet's best AURC. Confidence elicitation methods therefore
cannot be treated as interchangeable measurements of one latent quantity.

\begin{figure}[ht]
\centering
\includegraphics[width=.88\textwidth]{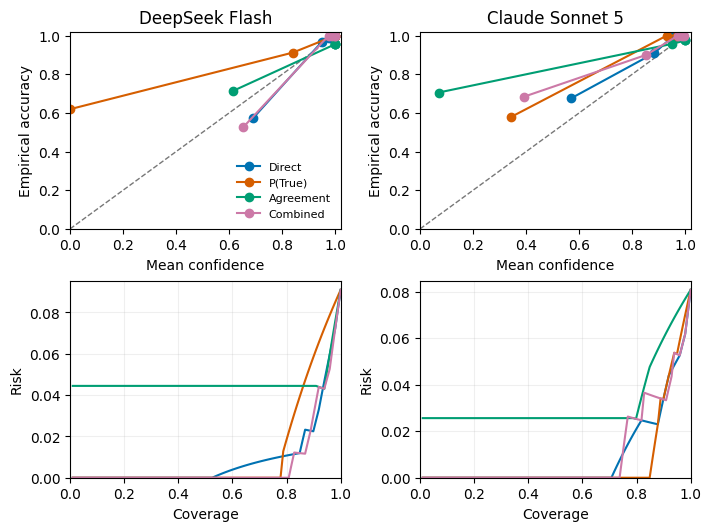}
\caption{Equal-mass calibration curves (top) and tie-aware risk and coverage
curves (bottom) on the audited TriviaQA samples. Agreement is coarse and often
assigns identical scores, while direct and post-hoc confidence provide more
useful rankings. Curves are descriptive because each model makes fewer than ten
errors.}
\label{fig:confidence-analysis}
\end{figure}

\subsection{Consensus repeats shared misconceptions}
Three-sample support is the weakest signal for both models, with AUROC .765 and
.790. More importantly, four of DeepSeek's nine audited errors and two of
Sonnet's eight errors receive unanimous support from all three additional
generations. Examples include an incorrect boxer attribution and an incorrect
historical ambulance origin. Independent decoding varies wording, but often
does not vary the underlying misconception.

The fixed interpolation improves DeepSeek's Brier, AUROC, and AURC point
estimates but slightly worsens its ECE; it worsens most Sonnet metrics. Paired
bootstrap intervals for the change all cross zero. For DeepSeek, the 95\%
intervals are $[-.0213,.0155]$ for Brier,
$[-.0289,.0857]$ for AUROC, and $[-.0090,.0026]$ for AURC; corresponding Sonnet
intervals are $[-.0050,.0450]$, $[-.0886,.0625]$, and $[-.0052,.0077]$.
Consequently, our data do not support a universal benefit from naive averaging.

\subsection{Equivalent prompts produce meaningfully different confidence}
Table~\ref{tab:prompt-robustness} compares re-elicited scores for the same fixed
answers. Mean absolute changes from the original $P(\mathrm{True})$ range from
.043 to .084. Between 4.0\% and 9.1\% of answers cross a confidence threshold of
.8 solely because of the prompt, and 4.0\% to 9.1\% change by at least .2. The
direction is not consistent: the paraphrase improves DeepSeek AUROC from .920
to .944, whereas both variants reduce Sonnet AUROC from .959 to .927 and .909.
Prompt choice is therefore part of the measurement procedure, not an innocuous
formatting detail.

\begin{table}[ht]
\caption{Sensitivity relative to the original post-hoc $P(\mathrm{True})$ on
fixed answers. MAE is mean absolute score change; flips cross the .8 threshold.}
\label{tab:prompt-robustness}
\centering
\small
\begin{tabular}{llrrr}
\toprule
Model & Re-elicitation & MAE & $|\Delta|\geq.2$ & .8 flips \\
\midrule
DeepSeek & Paraphrase & .084 & 9.1\% & 9.1\% \\
DeepSeek & Reversed scale & .043 & 4.0\% & 4.0\% \\
Sonnet & Paraphrase & .053 & 6.1\% & 4.0\% \\
Sonnet & Reversed scale & .066 & 7.1\% & 6.1\% \\
\bottomrule
\end{tabular}
\end{table}

\subsection{Claim-level confidence remains high on factual errors}
The biography audit labels 83 claims supported, five contradicted, and 12 as
insufficient evidence. Among the 88 decidable claims, mean confidence is .919
for supported claims and .830 for contradicted claims, with AUROC .884. One of
the five contradicted claims receives confidence at least .90. Because there are
only five negative claims, this result is descriptive, not a precise estimate of
long-form calibration. It nevertheless illustrates
the central failure mode: attaching a confidence to every claim does not ensure
that false claims receive low confidence.

\subsection{Benchmark auditing changes the conclusion}
Exact-match scoring initially marks several clearly correct variants as wrong,
including ``Hazell'' versus ``James Hazell'' and ``Dolenz, The Monkees'' versus
``Mickey Dolenz (The Monkees).'' One gold answer also lists ten Grand Slam titles
for Jimmy Connors, while the generated answer of eight is correct. Correcting
these cases raises measured accuracy by four points for DeepSeek and three for
Sonnet, and substantially improves apparent calibration. Confidence research is
therefore unusually sensitive to label quality: a mislabeled high-confidence
correct answer appears exactly like the overconfidence phenomenon being studied.

\section{Limitations}
Verbalized confidence may reflect learned rhetoric rather than internal
uncertainty. Automatic factuality judges may also introduce correlated errors.
Post-hoc evaluation and repeated sampling require additional model calls, so the
signals are training-free but not cost-free. Calibration may transfer poorly to
specialized domains, and confidence displays can cause automation bias: users
may over-trust a precise number even when aggregate calibration is good.

Our factuality setting does not cover subjective advice, creative writing, or
claims whose truth changes over time. Finally, proprietary model updates may
alter both answer quality and calibration; all conclusions are conditional on
the recorded model snapshots and evaluation dates. The short-answer study has
only 99 retained questions per model, and the long-form audit has only five
evidence-contradicted claims. Our confidence intervals and cautious framing
reflect this limited statistical power. Sonnet 5 also rejects explicit
temperature control on the evaluated endpoint, so its repeated samples use the
provider default rather than a matched temperature.

\section{Broader Impact}
Reliable uncertainty communication could help users identify claims requiring
verification and reduce confident hallucinations in high-stakes workflows.
Conversely, displaying confidence can create false reassurance, especially when
evaluation domains do not match deployment. The method should therefore be used
to allocate verification effort, not as a guarantee of truth. We do not evaluate
medical or legal deployment decisions, and low-confidence filtering must not be
used to systematically suppress information about underrepresented groups.

\section{Conclusion}
Our analysis finds that language models can emit useful confidence alongside
their answers without training, but different confidence signals are not
interchangeable. Direct verbalization is competitive with more expensive
alternatives, post-hoc self-evaluation varies by model, and repeated sampling can
confidently reproduce a shared error. A fixed mixture consequently fails to
generalize across the two evaluated model families. Even for a fixed answer,
equivalent confidence prompts can change scores enough to alter threshold-based
decisions. These results favor using confidence as a trigger for external
verification rather than treating a single elicitation or model agreement as
evidence of truth. Larger, independently adjudicated long-form studies are
needed before deploying claim-level confidence as a control signal.

\bibliography{references}
\bibliographystyle{iclr2027_conference}
\end{document}